\documentclass{article}
\usepackage[a4paper,margin=2.5cm]{geometry}
\usepackage{algorithm}
\usepackage{algpseudocode}
\usepackage{hyperref}
\usepackage{booktabs}
\usepackage{graphicx}
\usepackage{amsmath}

\newtheorem{example}{Example}[section]

\begin{document}
	\thispagestyle{plain}
	\vspace*{-2cm}
	\vspace*{0.5cm}
	
	\begin{center}
		{\Large\bf Sentence Splitter: Uncovering Latent Factual Structure for Self-Supervised Learning}
		\vspace*{0.5cm}
		
		{\small Ahmad Pouramini}\index{Pouramini, Ahmad}\footnote{speaker}, 
		{\small Mahsa Afsharizadeh}\index{Afsharizadeh, Mahsa}\\
		{\small Department of Computer Engineering, Sirjan University of Technology}\\[2mm]
		
	\end{center}
	
	\vspace*{0.5cm}
	
	\hspace{-1cm}\rule{\textwidth}{0.2mm}

\begin{abstract}
	This paper introduces \emph{Sentence Splitter}, a self-supervised framework built upon a T5-based encoder--decoder architecture for uncovering the latent factual structure of natural language sentences. The proposed method identifies the semantic boundary between a descriptive prefix (\emph{head}) and its factual completion (\emph{tail}) by formulating sentence splitting as a discrete segmentation problem, where a sentence of length $N$ admits $N$ possible split points but only one recovers the intended head--tail structure. Rather than explicitly searching over all candidate boundaries, the model learns to recover the factual completion through probabilistic sequence generation. To eliminate the need for manual annotation, symbolic head--tail pairs are first verbalized into natural-language templates that provide supervision for training the Sentence Splitter. The trained splitter is then applied to raw text to extract aligned prefix--tail pairs, which are subsequently used to train a generative model that proposes additional plausible completions through a lightweight bootstrapping process. This unified pipeline provides a scalable and structure-aware approach to constructing self-supervised training data while bridging symbolic knowledge and natural language. Experiments on both structured and naturally occurring text demonstrate that the proposed splitter generalizes beyond synthetic templates and that the resulting structure-aware supervision consistently improves downstream performance on knowledge graph completion and commonsense question answering, highlighting the effectiveness of recovering latent factual structure for knowledge-centric NLP.
\end{abstract}

\paragraph{Keywords:}
Sentence Splitting; Self-Supervised Learning; Natural Language Processing; Semantic Decomposition

\hspace{-1cm}\rule{\textwidth}{0.2mm}

\section{Introduction}

Natural language often encodes factual structure implicitly: a descriptive prefix (the \emph{head}) introduces an entity or situation, followed by a completion segment (the \emph{tail}) that supplies the key factual value. In a sentence of length $N$, any of the $N$ token boundaries may serve as the correct split between these two components. Identifying this boundary forms a discrete segmentation problem, one instance of the broader class of NLP tasks that require recovering latent structure inside text.

This work introduces a \emph{Sentence Splitter}, a model trained to recover the \textit{informative} completion segment of naturally phrased sentences while identifying the accompanying \textit{descriptive} prefix as contextual input. Unlike rule-based or parser-dependent pipelines, the splitter resolves segmentation implicitly through probabilistic decoding, mapping a sentence directly to its latent tail without explicitly enumerating candidate boundaries.

We formulate the problem as sentence splitting rather than general span masking. While factual information can, in principle, occur anywhere in a sentence, focusing on suffix completions provides a strong inductive bias aligned with knowledge graph verbalizations and enables a simple, parser-free learning objective.

The resulting prefix--tail pairs form a scalable, semantically aligned supervision signal for structure-aware learning. In addition to pre-training, these pairs naturally support an adaptation or fine-tuning phase, where models are trained to predict the tail given its prefix. This provides a lightweight structured resource for query answering and knowledge retrieval without requiring explicit knowledge graph construction.

The main contributions of this work are summarized as follows:

\begin{itemize} 
	\item \textbf{Discrete sentence splitting.}
	We formulate factual sentence splitting as a discrete decision problem over candidate segmentation boundaries and solve it through probabilistic sequence generation, avoiding explicit combinatorial search.

	\item \textbf{Symbolic self-supervision.}
	We introduce a self-supervised training strategy that automatically generates supervision by verbalizing symbolic head--tail pairs into natural-language sentences, enabling scalable learning from structured knowledge without manual annotation.
	
	\item 
	 \textbf{Bootstrapped structure-aware learning.}
	We propose a lightweight bootstrapping framework in which the Sentence Splitter extracts prefix--tail pairs from raw text and a generative model expands them with plausible factual completions, producing richer structure-aware supervision for downstream learning.
\end{itemize}

Unlike span-corruption objectives, which mask randomly selected spans, or segmentation approaches that rely on annotated boundaries, the proposed framework learns semantically meaningful factual completions from automatically generated supervision while remaining independent of external parsers.

The remainder of the paper is organized as follows. Section~\ref{sec:related} reviews related work. Section~\ref{sec:method} presents the proposed Sentence Splitter and the overall framework. Section~\ref{sec:experiments} reports the experimental evaluation, and Section~\ref{sec:discussion} discusses the results, limitations, and future research directions.

\section{Related Work}
\label{sec:related}

Research on sentence segmentation and decomposition has a long history. Classical syntactic parsers---from early probabilistic grammars to modern dependency parsers---expose hierarchical spans and have been used for segmentation through syntactic scoring \cite{favre2008segmentation}. Sentence simplification methods treat segmentation as partitioning a complex sentence into coherent clauses \cite{narayan2017split}. Other work models segmentation as a latent combinatorial variable, such as dynamic-programming-based marginalization \cite{wang2017segmental} or Segmental RNNs with semi-Markov structure \cite{kong2016srnn}. These approaches show that segmentation can be learned implicitly, but they typically require annotated parses or supervised clause boundaries.

Self-supervised pretraining has largely relied on corruption-based objectives such as masked language modeling \cite{bert,t5} or autoregressive next-token prediction \cite{gpt3}. Span-corruption variants \cite{t5,ul2} introduce contiguity but select spans randomly, without modeling interpretable semantic units. Structure-aware denoising approaches attempt to bias models toward linguistic structure by masking constituents \cite{constituentmasking} or dependency subtrees \cite{depbert}, or by corrupting semantic roles \cite{semslot}. These methods depend on external parsers, making them brittle in noisy or domain-shifted settings.

A parallel line of work links symbolic knowledge with natural language. Knowledge graph completion models generate or rank textualized triples \cite{comet2020,kgbert,kepler}, often assuming that the head--tail segmentation is already given. Our work instead treats the segmentation itself as the prediction target, discovering factual completion segments in free-form text.

Bootstrapped data generation techniques such as Self-Instruct \cite{selfinstruct}, Evol-Instruct \cite{evolinstruct}, and STaR \cite{star} refine training sets using model-generated examples. Knowledge-intensive pipelines often rely on stronger teacher models to produce high-quality explanations \cite{orca,flan}. Our approach shares the spirit of iterative refinement but differs in its structural focus: the Sentence Splitter learns to induce prefix--tail pairs that can be expanded by a generator into a richer factual dataset.

Across these threads, the Sentence Splitter introduces a distinct, scalable objective centered on recovering factual completions in natural text. It bridges structural pretraining, latent segmentation, and knowledge-centric generation while avoiding external parsing or manual annotation.

The next section presents the Sentence Splitter and the objective used to learn semantic head–tail boundaries.

\section{Method}
\label{sec:method}
\subsection{Sentence Splitter}

A natural sentence of length $N$ admits $N$ possible prefix--tail split points.  
Each token boundary is a candidate location where a factual completion may begin,  
yielding a discrete search space that grows linearly with sentence length.

Rather than explicitly enumerating or scoring all $N$ candidates, our approach uses  
a neural model to approximate this combinatorial decision in a single forward pass.  
Given a naturally phrased sentence $S$ describing a factual pair $(h,t)$, the  
Sentence Splitter predicts only the tail segment $t$.  
The remainder of the sentence becomes the prefix $p$, producing aligned $(p,t)$  
training pairs without any manual annotation. A valid split is one in which the tail corresponds to a minimal, contiguous span that can substitute for the tail element of an underlying symbolic fact, while preserving the original head context. Splits that are syntactically valid but fail to isolate this factual completion are considered incorrect.

For illustration, consider the example in Example~\ref{ex:valid-split}.  
Only one split corresponds to the correct factual completion; other token-level  
splits remain syntactically valid but semantically incorrect.

\begin{example}\label{ex:valid-split}
	Let the input sentence be:  
	\begin{quote}
		\emph{``To drive a car one needs to obtain a license.''}
	\end{quote}
	
	A valid split that corresponds to the underlying factual pair is:
	\[
	\underbrace{\text{to drive a car one needs}}_{p}
	\;\mid\;
	\underbrace{\text{to obtain a license}}_{t}.
	\]
	
	In contrast, the following segmentation is unacceptable because it doesn't isolate the true factual completion:
	\[
	\underbrace{\text{to drive}}_{p}
	\;\mid\;
	\underbrace{\text{a car one needs to obtain a license}}_{t}.
	\]

\end{example}

\subsection{Learning Objective}

Structured knowledge in the form of head--tail pairs is first verbalized into natural sentences using simple templates. The model then learns an approximation to the mapping
\begin{equation}\label{eq:map}
	f_\theta : S \rightarrow t,
\end{equation}
where $S$ is a complete input sentence and $t$ is the corresponding factual completion (tail) used as the supervision signal during training.

Although a sentence of length $N$ admits $N$ possible split points, the model does not explicitly enumerate or score these candidates. Instead, the decoder implicitly approximates the optimal completion by generating the sequence
\[
t^\ast = \arg\max_t P_\theta(t\,|\,S),
\]
thereby replacing explicit combinatorial search with probabilistic sequence generation.

The current formulation assumes that each input sentence is associated with a single dominant factual completion. Although a sentence may contain multiple factual fragments, the symbolic head--tail pairs used for supervision provide exactly one target tail for each training instance. Consequently, the decoder predicts the single completion that maximizes $P_\theta(t\,|\,S)$ under the learned model. Extending the framework to jointly identify multiple factual completions within a sentence is left as future work.

The decoder itself is not constrained to generate a span occurring in the input sentence. Consequently, the predicted output $\hat{t}=f_\theta(S)$ may occasionally differ from any contiguous segment of $S$. During the extraction stage, however, a prediction is accepted only if it exactly matches a \emph{contiguous token span} of the input sentence. Predictions that do not satisfy this condition are discarded.

For every accepted prediction, the matched span is removed from the sentence to produce the corresponding prefix $p$, yielding an aligned prefix--tail pair
\[
S = p \oplus \hat{t}, \qquad (p,\hat{t}) \in \mathcal{D}_{\text{split}},
\]
where $\oplus$ denotes concatenation of the extracted prefix and tail.

A minimal version of this procedure is summarized in Algorithm~\ref{alg-basic}.

\begin{algorithm}[!ht]
	\caption{Sentence Splitter (Single-Sentence Case)}
	\begin{algorithmic}[1]
		\Require sentence $S$, model $f_\theta$
		\State $\hat{t} \gets f_\theta(S)$ \Comment{Implicit maximization of $P_\theta(t \mid S)$}
		\If{$\hat{t}$ exactly matches a contiguous token span of $S$}
		\State $p \gets S$ with the matched span removed
		\State \Return $(p,\hat{t})$
		\Else
		\State \Return discard
		\EndIf
	\end{algorithmic}
	\label{alg-basic}
\end{algorithm}

This core mechanism serves as the foundation for the broader pipeline, where the extracted prefix--tail pairs are subsequently used for corpus-level structural extraction and generative data augmentation, as described in the following section.

\subsection{Pipeline}

\begin{figure}[t]
	\centering
	\includegraphics[width=\linewidth]{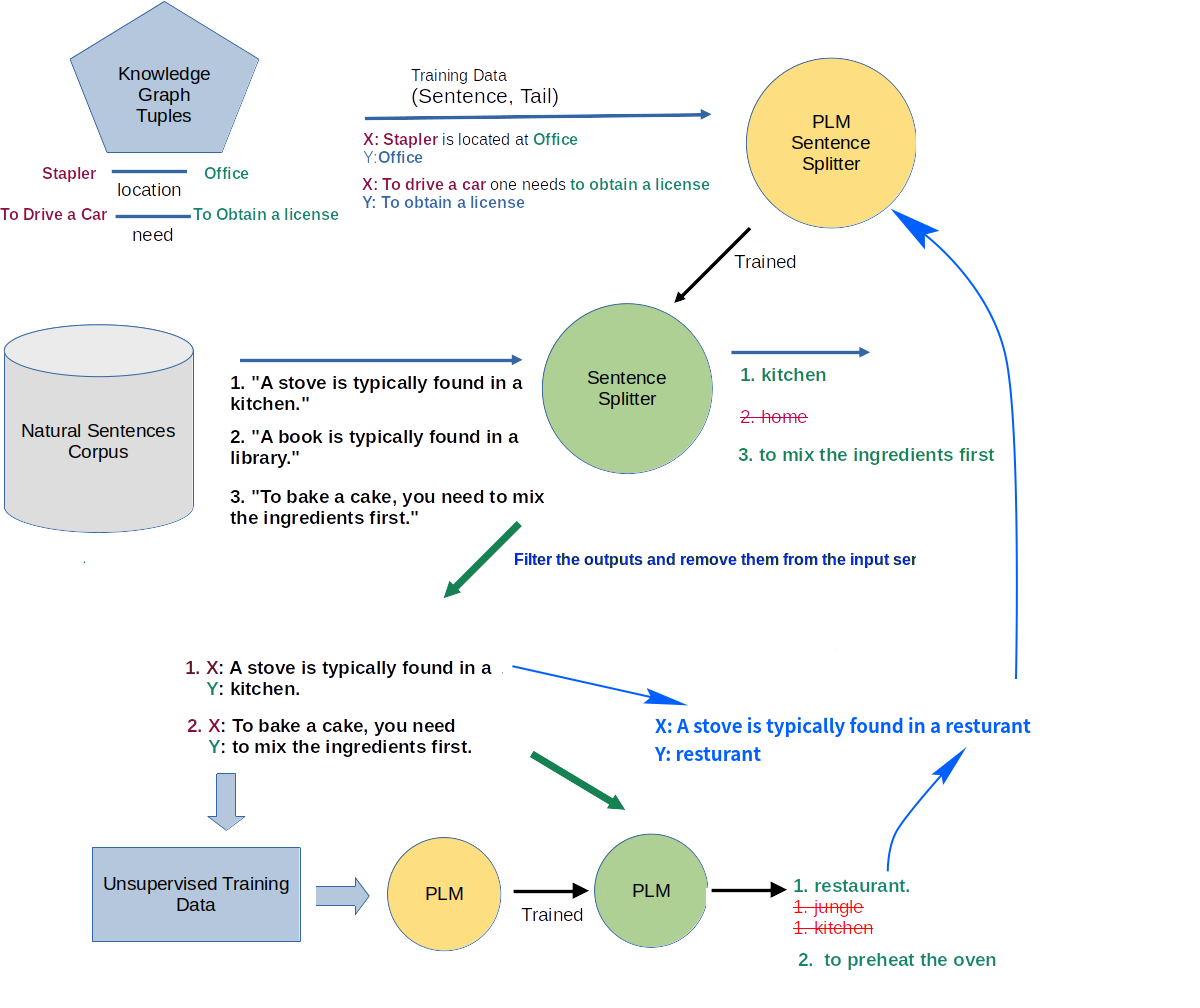}
	\caption{The full pipeline for training a sentence-splitter and expanding its supervised data using a trained generative model.}
	\label{fig:pipeline}
\end{figure}

The overall workflow is illustrated in Figure~\ref{fig:pipeline}. The process begins with symbolic knowledge graph tuples $(h,r,t)$, which are verbalized into natural sentences through lightweight templates. Each verbalized sentence is paired with its known tail $t$, forming the initial supervised dataset
\[
\mathcal{D}_{\text{sup}}=\{(S,t)\}.
\]

This dataset is used to fine-tune the \emph{Sentence Splitter} model $f_\theta^{\text{split}}$, whose task is to recover the correct tail segment inside a sentence. For a sentence of length $N$, there are $N$ possible token boundaries; the model learns to identify the one corresponding to the factual completion.

After this initial training, the splitter is applied to a large raw corpus $\mathcal{C}$. For each sentence $S\in\mathcal{C}$, the model predicts a candidate tail segment by implicitly solving

\[
t^\ast=\arg\max_t P_\theta^{\text{split}}(t\,|\,S),
\]

which replaces explicit combinatorial search over token boundaries with probabilistic sequence generation. If the predicted tail exactly matches a contiguous token span of $S$, it is extracted and removed to yield a prefix--tail pair $(p,t^\ast)$, forming the dataset

\[
\mathcal{D}_{\text{split}}
=
\{(p,t^\ast)\}.
\]

A second model, the \emph{Generator} PLM $f_\theta^{\text{gen}}$, is then fine-tuned on $\mathcal{D}_{\text{split}}$. Its role is to generate plausible alternative tails for a given prefix. For every $(p,t)\in\mathcal{D}_{\text{split}}$, the generator predicts

\[
\tilde t=f_\theta^{\text{gen}}(p),
\]

from which a reconstructed sentence

\[
S_{\text{new}}=p\oplus\tilde t
\]

is obtained.

Before entering the augmented training set, each reconstructed sentence is passed through an optional validation stage whose purpose is to eliminate obviously invalid or low-quality generations. This validation stage is independent of the proposed sentence-splitting framework and may be instantiated using application-specific quality criteria.

The accepted examples form the augmented dataset

\[
\mathcal{D}_{\text{aug}}
=
\{(S_{\text{new}},\tilde t)\}.
\]

The augmented dataset is merged back into the supervised pool

\[
\mathcal{D}_{\text{sup}}
\gets
\mathcal{D}_{\text{sup}}
\cup
\mathcal{D}_{\text{aug}},
\]

and the splitter $f_\theta^{\text{split}}$ is retrained on this enlarged dataset. This establishes a mild bootstrapping loop: the splitter extracts structural regularities from raw text, the generator proposes new tails consistent with those structures, and the splitter benefits from the enriched training signal.

\[
\mathcal{D}_{\text{sup}}
\xrightarrow{\mathrm{train}}
f_\theta^{\text{split}}
\xrightarrow{\mathrm{split}}
\mathcal{D}_{\text{split}}
\xrightarrow{\mathrm{train}}
f_\theta^{\text{gen}}
\xrightarrow{\mathrm{generate}}
\mathcal{D}_{\text{aug}}
\xrightarrow{\mathrm{merge}}
\mathcal{D}_{\text{sup}}.
\]

\subsection{Validation of Generated Sentences}

The validation stage shown in Figure~\ref{fig:pipeline} serves to prevent low-quality generated examples from being incorporated into the augmented training set. Since the primary contribution of this work is the proposed sentence-splitting framework rather than sentence-quality estimation, the validation module is intentionally kept independent of the proposed method.

Depending on the target application, this module may employ lightweight lexical or statistical criteria, such as minimum tail length, removal of duplicated or repetitive generations, language-model fluency or perplexity thresholds, or semantic similarity measures between the generated and original sentences. Such criteria are implementation dependent and may be selected to balance generation quality and data diversity.

The experiments presented in this paper focus on evaluating the effectiveness of the proposed sentence-splitting framework. Consequently, no specific validation strategy is optimized or compared, leaving the design of task-specific validation modules as an interesting direction for future work.

A complete specification of the pipeline, including its iterative bootstrapping loop, is provided in Algorithm~\ref{alg:full_pipeline}.

\begin{algorithm}[!ht]
	\caption{Full Structure-Aware Self-Supervision Pipeline (with Bootstrapped Loop)}
	\label{alg:full_pipeline}
	\begin{algorithmic}[1]
		
		\Require Knowledge graph tuples $\mathcal{K}$, raw corpus $\mathcal{C}$,
		splitter PLM $f_\theta^{\text{split}}$, generator PLM $f_\theta^{\text{gen}}$,
		max iterations $T$
		
		\Statex \textbf{Stage 1: Initial Supervised Dataset from KG}
		
		\State $\mathcal{D}_{\text{sup}}\gets\emptyset$
		\ForAll{$(h,r,t)\in\mathcal{K}$}
		\State $S\gets\text{textualize}(h,r,t)$
		\State add $(S,t)$ to $\mathcal{D}_{\text{sup}}$
		\EndFor
		
		\Statex \textbf{Stage 2: Initial Training}
		
		\State Fine-tune $f_\theta^{\text{split}}$ on $\mathcal{D}_{\text{sup}}$
		
		\Statex \textbf{Stage 3: Bootstrapped Loop}
		
		\For{$i=1$ to $T$}
		
		\Statex \textbf{(a) Split corpus using the splitter model}
		
		\State $\mathcal{D}_{\text{split}}\gets\emptyset$
		
		\ForAll{$S\in\mathcal{C}$}
		\State $\hat t\gets f_\theta^{\text{split}}(S)$
		\If{$\hat t$ exactly matches a contiguous token span of $S$}
		\State $p\gets S$ with the matched span removed
		\State add $(p,\hat t)$ to $\mathcal{D}_{\text{split}}$
		\EndIf
		\EndFor
		
		\State Fine-tune $f_\theta^{\text{gen}}$ on $\mathcal{D}_{\text{split}}$
		
		\Statex \textbf{(b) Apply the generator model to prefixes}
		
		\State $\mathcal{D}_{\text{aug}}\gets\emptyset$
		
		\ForAll{$(p,t)\in\mathcal{D}_{\text{split}}$}
		\State $\tilde t\gets f_\theta^{\text{gen}}(p)$
		\State $S\gets p\oplus\tilde t$
		\If{$S$ passes the validation module}
		\State add $(S,\tilde t)$ to $\mathcal{D}_{\text{aug}}$
		\EndIf
		\EndFor
		
		\Statex \textbf{(c) Merge and retrain both models}
		
		\State $\mathcal{D}_{\text{sup}}\gets
		\mathcal{D}_{\text{sup}}\cup
		\mathcal{D}_{\text{aug}}$
		
		\State Fine-tune $f_\theta^{\text{split}}$ on $\mathcal{D}_{\text{sup}}$
		
		\EndFor
		
		\State \Return $\mathcal{D}_{\text{split}}$
		
	\end{algorithmic}
\end{algorithm}

\section{Experiments}
\label{sec:experiments}

This section evaluates the proposed structure-aware self-supervision pipeline. The experiments focus on knowledge graph completion and commonsense reasoning tasks, using well-established benchmarks and pretrained encoder--decoder models. The goal is to measure whether the Sentence Splitter and the subsequent augmentation loop produce training signals that improve downstream performance.

\subsection{Datasets}

\paragraph{ATOMIC2020.}
We conduct experiments on a subset of relations from ATOMIC2020 \cite{comet2020}, an expanded version of the ATOMIC commonsense knowledge graph \cite{atomic}. ATOMIC2020 contains approximately 1.3M triplets of the form $(h,r,t)$, covering $23$ relation types. These relations are grouped into three broad categories: (1) social interactions and intents, (2) physical entities and affordances, and (3) general event dynamics such as causal or temporal relations. Some of the physical-entity relations incorporate content from ConceptNet \cite{conceptnet}, producing a comprehensive commonsense resource.

Each triplet can be verbalized as a natural sentence using lightweight templates. Relations centered on social situations naturally correspond to conditional statements. For example:
\[
\text{If PersonX cooks, he intends to satisfy hunger.}
\]
corresponds to the relation \texttt{xIntent}. A single head may be associated with multiple tails for the same relation, reflecting diverse plausible consequences.

\paragraph{CommonsenseQA.}
For downstream evaluation, we use CommonsenseQA, a multiple-choice benchmark requiring reasoning about everyday physical and social situations. The task measures a model’s ability to apply commonsense knowledge beyond the explicit wording of the question.

\paragraph{OMCS corpus.}
To supply raw, naturally phrased training material, we use the Open Mind Common Sense (OMCS) corpus \cite{omcs}. OMCS consists of more than 700{,}000 crowdsourced sentences expressing ordinary commonsense knowledge. Each sentence is accompanied by a numeric quality score. For training the splitter and generator models in a self-supervised manner, we select the top 8{,}000 highest-scoring sentences. These sentences are then segmented into prefix–tail pairs using the Sentence Splitter described in Section~\ref{sec:method}.

\subsection{Models}

We use T5 \cite{t5}, an encoder–decoder Transformer well-suited for prefix-to-sequence prediction. T5-base is pretrained on the C4 corpus with a denoising objective, followed by an additional 100k steps of causal language modeling. Both T5-base and T5-large are further trained in a multitask setting that mixes supervised objectives with self-supervised denoising. These models have been widely used to generate tails in commonsense knowledge graphs such as ATOMIC, which forms a natural foundation for our pipeline \cite{comet2020,analyzingCS,rainbow,unified}.

\subsection{Sentence Splitter Evaluation}

To evaluate the Sentence Splitter independently of the augmentation loop, we first assess its ability to recover factual completions from structured sentences generated from knowledge graph tuples. Following the training protocol, the model is evaluated on 1{,}000 verbalized ATOMIC relation instances using exact-match accuracy. A prediction is counted as correct only if the predicted tail exactly matches the gold tail and appears as a contiguous token span of the input sentence.

Since these verbalizations closely resemble the synthetic training data, we additionally evaluate the model on naturally occurring text to assess its ability to generalize beyond templated sentences. We manually annotate the correct prefix--tail boundary for a randomly selected set of 100 sentences from the OMCS corpus and evaluate the splitter using the same exact-match criterion.

The results are summarized in Table~\ref{table:splitter}. As expected, performance is highest on the structured ATOMIC verbalizations, where the splitter achieves an exact-match accuracy of 96.0\%. More importantly, the model attains an accuracy of 84.0\% on manually annotated OMCS sentences, demonstrating that the learned segmentation strategy transfers reasonably well to naturally occurring text despite being trained primarily on templated knowledge-graph verbalizations. The remaining errors are largely attributable to the greater syntactic variability and linguistic complexity present in free-form language. These results support the central hypothesis of this work: supervision obtained from symbolic knowledge can be used to learn a sentence-splitting strategy that generalizes beyond synthetic templates and recovers meaningful factual completions in naturally written text.

\begin{table}[t]
	\centering
	\caption{Exact-match accuracy of the Sentence Splitter on structured and naturally occurring sentences.}
	\label{table:splitter}
	\begin{tabular}{lcc}
		\toprule
		Dataset & Samples & Exact Match (\%) \\
		\midrule
		ATOMIC verbalizations & 1,000 & 96.0 \\
		OMCS (manual annotation) & 100 & 84.0 \\
		\bottomrule
	\end{tabular}
\end{table}

\subsection{Downstream Evaluation}

We examine how structure-aware pretraining influences two downstream tasks:

\begin{itemize}
	\item \emph{CommonsenseQA}, evaluated using answer-selection accuracy.
	\item \emph{ATOMIC2020 completion}, evaluated using ROUGE scores \cite{rouge}, following the evaluation protocol of \cite{comet2020}.
\end{itemize}

All models are pretrained on the same pool of 3{,}000 sentences under three configurations:

\begin{itemize}
	\item (i) standard masked language modeling (MLM),
	\item (ii) MLM augmented with splitter-based self-supervision,
	\item (iii) the full pipeline, including one bootstrap iteration of generator-based augmentation.
\end{itemize}

For the standard MLM baseline, each sentence is converted into a denoising training example by randomly selecting a split boundary and replacing the resulting contiguous suffix with a special mask token in the encoder input, while the decoder is trained to reconstruct the masked span. Thus, the baseline is intentionally designed as a controlled comparison with the proposed method. Both approaches use the same training corpus, denoising objective, optimization procedure, and downstream fine-tuning protocol; they differ only in how the masked span is determined. The baseline selects the split boundary uniformly at random, whereas the proposed method predicts a semantically meaningful boundary using the Sentence Splitter.

To align the pretrained models with each downstream task, we perform light fine-tuning using 30 training examples per task, resulting in a few-shot evaluation setting. For ATOMIC2020, ROUGE scores are averaged across all 23 relation types.

To account for the effect of random initialization, each experiment is repeated using three different random seeds. Table~\ref{table:downstream} reports the mean performance together with the corresponding standard deviation.

\begin{table}[t]
	\centering
	\caption{Downstream task performance after structure-aware pretraining. Results are reported as mean $\pm$ standard deviation over three runs with different random seeds.}
	\label{table:downstream}
	\begin{tabular}{lcc}
		\toprule
		Pretraining setup & CommonsenseQA Acc. (\%) & ATOMIC ROUGE (\%) \\
		\midrule
		Standard MLM & $51.72 \pm 2.93$ & $20.65 \pm 1.23$ \\
		+ splitter (no augmentation) & $55.73 \pm 1.25$ & $25.14 \pm 1.31$ \\
		+ splitter + one bootstrap iteration & $57.31 \pm 0.85$ & $27.51 \pm 0.92$ \\
		\bottomrule
	\end{tabular}
\end{table}

The results demonstrate that the proposed sentence-splitting objective consistently improves downstream performance over the standard MLM baseline across different random initializations. While the baseline learns to reconstruct randomly selected contiguous spans, the proposed method replaces this random selection with semantically meaningful prefix--tail decompositions obtained by the Sentence Splitter. Incorporating the splitter alone yields substantial gains on both CommonsenseQA and ATOMIC2020, while a single bootstrap iteration provides an additional improvement. Moreover, the progressively smaller standard deviations suggest that the proposed structure-aware pretraining strategy produces not only higher average performance but also more stable learning behavior.

The present study intentionally evaluates only a single bootstrap iteration in order to isolate the contribution of the proposed structure-aware augmentation from the effects of repeated self-training. While additional iterations may further improve performance, they may also introduce semantic drift or error accumulation in the generated supervision signal. A systematic investigation of the trade-off between iterative improvement and error propagation is beyond the scope of this work and is therefore left for future research.

Overall, these findings indicate that identifying factual completions in natural sentences provides a more informative structured supervision signal than randomly masking contiguous spans.

\section{Discussion}
\label{sec:discussion}

The experimental results demonstrate that the proposed Sentence Splitter reliably identifies meaningful semantic boundaries in both structured and naturally occurring text. Its high exact-match accuracy on templated ATOMIC verbalizations confirms that the proposed discrete segmentation objective is readily learnable, while the evaluation on manually annotated OMCS sentences indicates that the learned segmentation strategy generalizes beyond synthetic templates to free-form language. Rather than relying on handcrafted rules, dependency parsers, or explicit boundary enumeration, the model implicitly learns to recover semantically meaningful factual completions through probabilistic sequence generation.

Because the supervision required by the framework consists only of symbolic head--tail pairs, the proposed approach naturally extends beyond knowledge graph verbalizations. Any domain containing latent factual completions---including encyclopedic articles, instructional documents, scientific literature, biomedical records, or legal documents---can potentially benefit from this learning paradigm. The lightweight bootstrapping loop further enhances this capability by allowing the generator to propose additional plausible factual completions, gradually enriching the diversity of structural patterns available for subsequent training.

The proposed framework therefore establishes a principled connection between symbolic knowledge, discrete sentence segmentation, and large-scale self-supervised learning. The consistent improvements observed on both CommonsenseQA and ATOMIC2020 suggest that recovering latent factual structure provides a complementary supervision signal beyond conventional masked language modeling, improving both downstream performance and training stability across different random initializations.

More broadly, the proposed framework suggests a general strategy for domain adaptation through structure-aware bootstrapping. Starting from a relatively small collection of domain-specific symbolic knowledge, such as a knowledge graph or ontology, the Sentence Splitter can progressively extract domain-relevant factual completions from large unlabeled corpora. These extracted examples can then be used to continually retrain both the splitter and the generator, allowing the supervision signal itself to evolve toward the target domain. Evaluating this adaptation strategy on specialized domains, such as biomedical and legal corpora, represents an important next step toward assessing the generalizability and practical applicability of the proposed framework.

Beyond domain adaptation, the proposed framework also opens new research directions for improving the bootstrapping process itself. In the current implementation, generated sentences are accepted using simple validation rules. A promising extension is to replace these fixed criteria with adaptive acceptance mechanisms that optimize the quality of generated supervision during training. For example, reinforcement learning could be employed to dynamically adjust sentence acceptance according to downstream task performance, confidence estimates, or semantic consistency, allowing the system to progressively refine its own supervision policy.

Finally, the proposed framework raises broader questions regarding the dynamics of iterative self-supervision. In the present work, only a single augmentation iteration is evaluated in order to isolate the contribution of structure-aware supervision. However, repeated bootstrapping may introduce semantic drift, error accumulation, or memorization effects that are not yet fully understood. Future work will therefore investigate the stability of long-term bootstrapping through confidence-aware filtering, adaptive sample selection, and quantitative measures of semantic consistency. More generally, understanding when iterative self-supervision improves generalization and when it leads to degradation or collapse represents an important direction for developing reliable, scalable, and structure-aware learning systems.

\section{Conclusion}

This paper introduced Sentence Splitting, a T5-based encoder--decoder framework that formulates factual sentence decomposition as a discrete segmentation problem. By automatically generating supervision from symbolic knowledge graph tuples, the proposed approach learns semantically meaningful prefix--tail decompositions without requiring manually annotated split boundaries. Experimental results demonstrate that the learned splitter generalizes from templated knowledge graph verbalizations to naturally occurring text and that the resulting structure-aware supervision consistently improves downstream performance on knowledge graph completion and commonsense question answering.

A limitation of the current framework is that it assumes the factual completion forms a single contiguous suffix of the sentence. While this assumption substantially simplifies the learning objective and aligns naturally with knowledge graph verbalizations, many naturally occurring sentences express relevant factual information in the middle of a sentence, through multiple spans, or via non-contiguous expressions. Consequently, the current model cannot recover all possible forms of latent factual structure.

Future work will therefore focus on extending the proposed framework to more general segmentation settings capable of predicting arbitrary or multiple factual spans while preserving the scalability and interpretability of the structure-aware learning paradigm. As discussed in the previous section, we also envision broader extensions toward domain-specific adaptation and more adaptive bootstrapping strategies, further advancing the integration of symbolic knowledge and self-supervised learning for robust knowledge-aware language models.


	% ...

\end{document}